%% file: main.tex
\documentclass[10pt,twocolumn,letterpaper]{article}
\usepackage{graphicx}
\usepackage{amsmath}
\usepackage{amssymb}
\usepackage{booktabs}
\usepackage[pagebackref,breaklinks,colorlinks]{hyperref}
\usepackage[capitalize]{cleveref}
\crefname{section}{Sec.}{Secs.}
\Crefname{section}{Section}{Sections}
\Crefname{table}{Table}{Tables}
\crefname{table}{Tab.}{Tabs.}
\def\wacvPaperID{*****} 
\def\confName{WACV}
\input{regpkg}

\input{definitions} 
\begin{document}
\title{Conditional
Synthetic Live and Spoof Fingerprint Generation}
\newcommand{\clrksn}{{*}}
\newcommand{\afrl}{{$^{\ddagger}$}}
\renewcommand*{\Authsep}{\ }
\renewcommand*{\Authand}{\ }
\renewcommand*{\Authands}{\ }
\author[\clrksn]{Syed Konain Abbas}
\author[\clrksn]{Sandip Purnapatra}
\author[\afrl]{M. G. Sarwar Murshed}
\author[\clrksn]{Conor Miller-Lynch} 
\author[\clrksn]{Lambert Igene} 
\author[\clrksn]{Soumyabrata Dey} 
\author[\clrksn]{Stephanie Schuckers} 
\author[\clrksn]{Faraz Hussain} 
\affil[\clrksn]{Clarkson University, Potsdam, NY, USA\authorcr {\tt \{abbas,purnaps,millerlc,sdey,sschucke,fhussain\}@clarkson.edu}\vspace{0.4em}}
\affil[\afrl]{University of Wisconsin-Green Bay, WI, USA\authorcr {\tt murshedm@uwgb.edu}\vspace{0.4em}}
\date{}
\maketitle
\begin{abstract}
Large fingerprint datasets, while important for training and evaluation, are time-consuming and expensive to collect and require strict privacy measures. Researchers are exploring the use of synthetic fingerprint data to address these issues. This paper presents a novel approach for generating synthetic fingerprint images (both spoof and live), addressing concerns related to privacy, cost, and accessibility in biometric data collection. Our approach utilizes conditional StyleGAN2-ADA and StyleGAN3 architectures to produce high-resolution synthetic live fingerprints, conditioned on specific finger identities (thumb through little finger). Additionally, we employ CycleGANs to translate these into realistic spoof fingerprints, simulating a variety of presentation attack materials (e.g., EcoFlex, Play-Doh). These synthetic spoof fingerprints are crucial for developing robust spoof detection systems. Through these generative models, we created two synthetic datasets (DB2 and DB3), each containing 1,500 fingerprint images of all ten fingers with multiple impressions per finger, and including corresponding spoofs in eight material types. The results indicate robust performance: our StyleGAN3 model achieves a Fréchet Inception Distance (FID) as low as 5, and the generated fingerprints achieve a True Accept Rate of 99.47 at a 0.01\% False Accept Rate. The StyleGAN2-ADA model achieved a TAR of 98.67\% at the same 0.01\% FAR. We assess fingerprint quality using standard metrics (NFIQ2, MINDTCT), and notably, matching experiments confirm strong privacy preservation, with no significant evidence of identity leakage, confirming the strong privacy-preserving properties of our synthetic datasets. 

\end{abstract}

\section{Introduction}\label{sec:intro}
\begin{figure}
 \centering
 \includegraphics[width=1\linewidth]{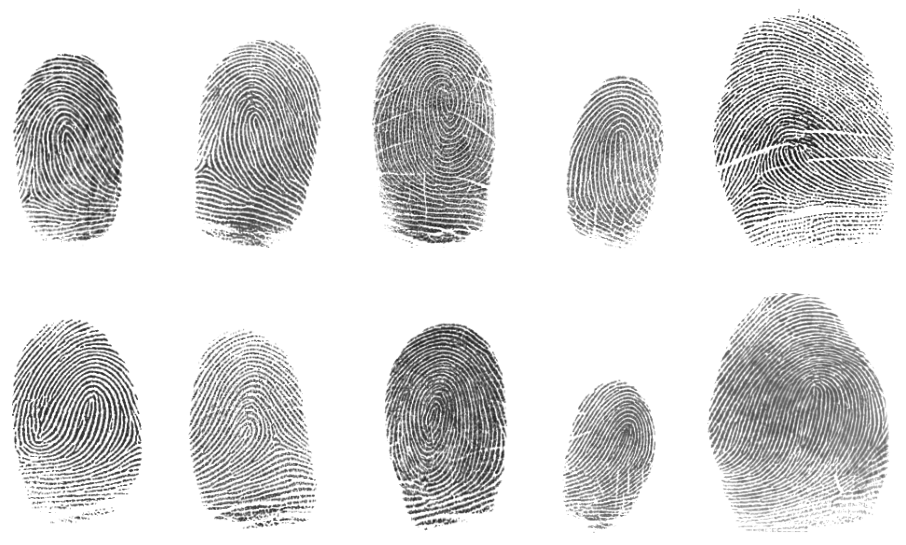} 
 \caption{Synthetic fingerprints, each sized $512 \times 512$ pixels, generated using our newly developed model described in Section ~\ref{subsubsec:stylegan3_training}, which leverages the architecture of StyleGAN3. The first row shows fingerprints for the Left-Index, Left-Middle, Left-Ring, Left-Little, and Left-Thumb, while the second row shows fingerprints for the Right-Index, Right-Middle, Right-Ring, Right-Little, and Right-Thumb.}
 \label{fig:gan3}
\end{figure}
\begin{figure}
 \centering
 \includegraphics[width=1\linewidth]{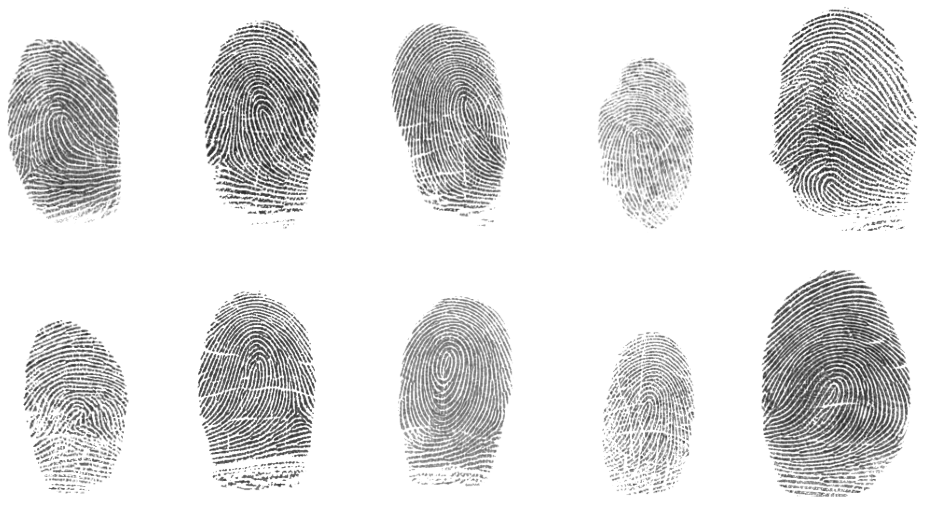} 
 \caption{Synthetic fingerprints, each sized $512 \times 512$ pixels, generated using our newly developed model described in Section ~\ref{subsubsec:stylegan2_training}, which leverages the architecture of StyleGAN2-ADA. The first row shows fingerprints for the Left-Index, Left-Middle, Left-Ring, Left-Little, and Left-Thumb, while the second row shows fingerprints for the Right-Index, Right-Middle, Right-Ring, Right-Little, and Right-Thumb.}
 \label{fig:gan2}
\end{figure}

Fingerprint identification systems rely heavily on large and diverse datasets for effective training and evaluation, especially when utilizing deep learning models such as Convolutional Neural Networks (CNNs). However, privacy concerns, data collection costs, and restrictive data sharing policies limit the availability of real biometric datasets. In response to these challenges, researchers are increasingly turning to synthetic fingerprint generation, which not only reduces costs associated with data collection but also alleviates privacy concerns by using artificially generated fingerprint images instead of real fingerprint data. \par
\begin{figure*}[t] 
 \centering
 \includegraphics[width=1\linewidth]{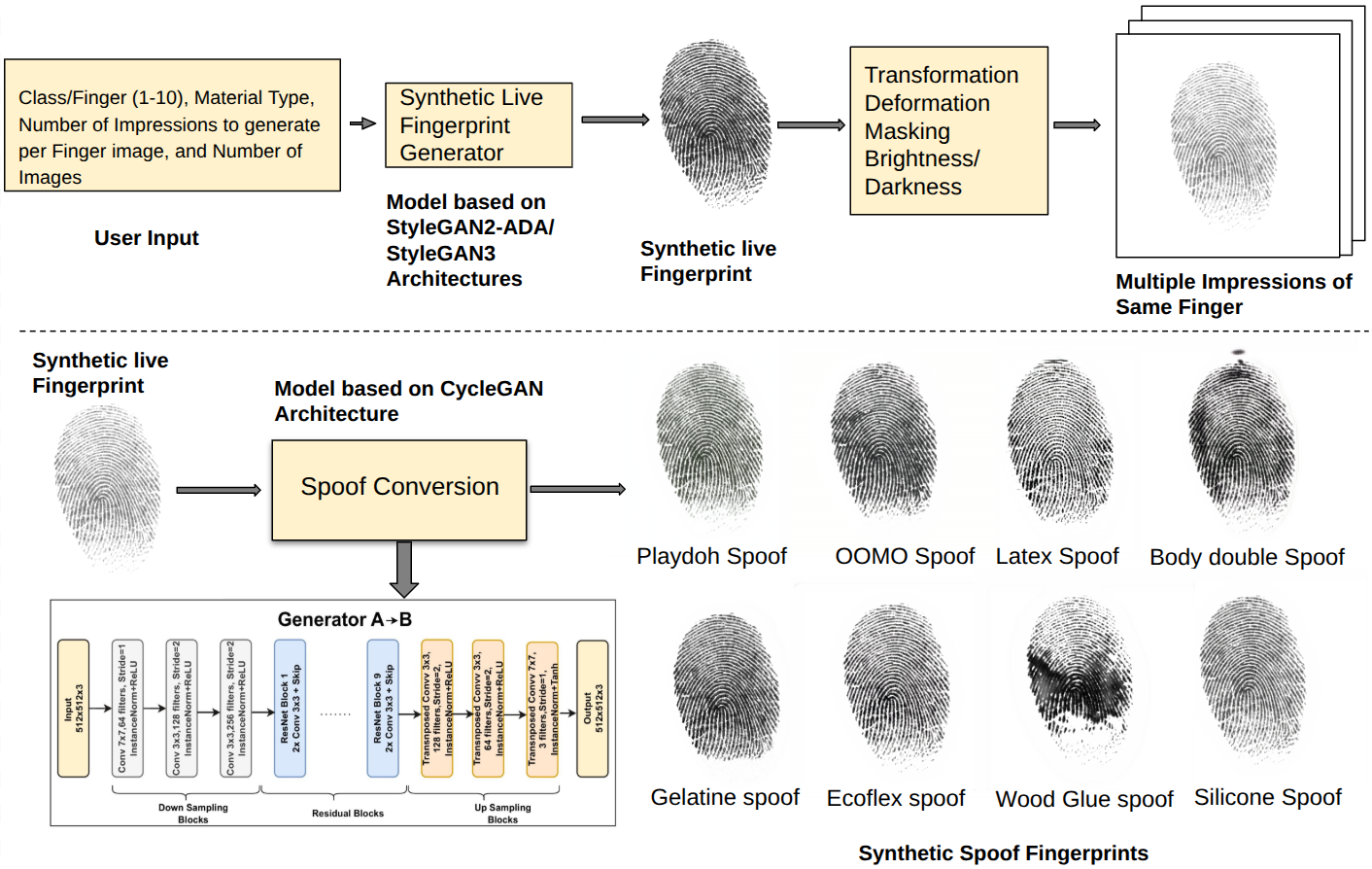} 
 \caption{The figure illustrates the overall architecture of our proposed fingerprint generation framework. Users are required to provide the finger/class number, material type (live or one of the eight spoof types), and specify the desired number of fingerprint samples along with the number of multiple impressions per finger.}
 \label{fig:overview}
\end{figure*}

Recent advances in generative models, particularly Generative Adversarial Networks (GANs), have made it possible to synthesize fingerprint images with high realism and uniqueness ~\cite{shoshan2024fpgan, wyzykowski2023synthetic, engelsma2022printsgan, grosz2022spoofgan, wyzykowski2021level, bahmani2021high}. These models can replicate the complex textures and ridge patterns characteristic of authentic fingerprints, making them ideal for training fingerprint recognition and spoof detection systems. 
The creation of spoof fingerprint datasets is vital for the advancement of fingerprint presentation attack detection (PAD) systems. As biometric technologies are increasingly deployed in security-critical applications, the threat of spoofing, where attackers use fake fingerprints to deceive recognition systems, has become a significant concern. Effective spoof detection models require large and diverse datasets that represent a wide range of spoof materials and fabrication techniques. However, collecting authentic spoof fingerprint data is both challenging and expensive, often constrained by ethical, privacy, and logistical issues. Synthetic spoof datasets offer a scalable and privacy-preserving alternative, enabling researchers to explore a wider range of attack scenarios, improve generalization to previously unseen spoofs, and evaluate PAD models under varied and realistic conditions. By generating high-quality spoof fingerprints, these datasets play a critical role in strengthening the security and robustness of modern biometric systems.

Several works have addressed synthetic fingerprint generation. For instance, the Clarkson Fingerprint Generator (CFG) \cite{bahmani2021high} employed a progressive growth approach, a GAN-based technique for generating high-fidelity plain impression fingerprints. However, it lacked the capability to generate a fingerprint based on a specific finger (1-10) and was unable to produce multiple impressions of the same fingerprint. PrintsGAN \cite{engelsma2022printsgan} introduced a model capable of generating realistic live fingerprint impressions and multiple samples per identity. However, it lacks the ability to generate fingerprints based on specific finger numbers (1–10) and does not address spoof fingerprint generation. CFG V2 \cite{bahmani2022deep} extends the Clarkson Fingerprint Generator by using a conditional GAN to generate fingerprints based on finger types and specific age groups (under 9, 9–14, adults). However, it does not support generating multiple impressions of the same fingerprint. SpoofGAN \cite{grosz2022spoofgan} extended this work by generating both live and spoof fingerprints of the same identity, using advanced texture rendering and distortion techniques, but it lacks class-based generation control, such as finger number conditioning.

In biometric applications, generating synthetic fingerprints across all finger classes is particularly important for comprehensive identification, such as in forensics or high-security access systems. Each finger contributes unique biometric features, and inter-finger variability must be considered for robust identity verification\cite{mishrareview}. Some datasets like NIST SD4 \cite{watson1992nist}, SD14\cite{watson1993nist14}, SD27\cite{garris2000nist27}, and Special Database 30 \cite{nist-db30} have been removed from public access due to increasing privacy regulations. The growing adoption of synthetic data in facial recognition, as seen in the Digiface \cite{bae2023digiface} and SFace \cite{boutros2022sface} datasets, highlights a broader shift toward privacy-conscious biometric development. This trend is now extending to fingerprint synthesis. To address the lack of publicly available fingerprint datasets, several studies have proposed algorithms for generating synthetic fingerprints. Since synthetic fingerprints are not linked to real individuals, they typically do not require data protection approvals under Institutional Review Board (IRB) guidelines \cite{engelsma2022printsgan}. As such, they offer a privacy-preserving solution ideal for public release and academic research.

Inspired by recent advancements, we introduce a conditional fingerprint generation framework based on StyleGAN2-ADA \cite{karras2020training} and StyleGAN3 \cite{karras2021alias}. Additionally, we employ CycleGAN\cite{zhu2017unpaired} to generate corresponding synthetic \emph{spoof} fingerprints that visually replicate specific attack materials such as Ecoflex, PlayDoh, Gelatine, and others. Our novel model is capable of producing high-resolution (512×512) live and spoof fingerprint images conditioned on specific finger classes (1–10), and it can also generate multiple impressions of the same finger. Two comprehensive datasets, DB2 and DB3, were constructed using our novel generative models: StyleGAN2-ADA and StyleGAN3, respectively. Each dataset comprises 1,500 live fingerprints alongside eight corresponding spoof variants. The eight spoofed images were synthesized using our custom-developed 8 CycleGAN architectures, specifically designed GAN models for spoof fingerprint image generation.

This paper makes the following \emph{novel contributions}:
\begin{itemize}
\item We developed conditional StyleGAN2-ADA, StyleGAN3, and CycleGAN-based models to generate high-fidelity synthetic live and spoof fingerprint images for each finger (1–10). To the best of our knowledge, this is the first fingerprint generation framework capable of producing both synthetic live and spoof fingerprints conditioned on specific finger numbers (1–10), as well as generating multiple impressions of the same finger.

\item We evaluate the quality and realism of our generated fingerprints using NIST NFIQ 2.0 \cite{nfiq20160} and MINDTCT \cite{tabassi2004nist}. Comparisons of synthetically generated fingerprint datasets with bona fide fingerprint dataset via VeriFinger \cite{neurotechnology_verifinger} show that our synthetically generated fingerprint datasets effectively preserve user privacy.


\end{itemize}

\section{Related Work}
Early approaches to synthetic fingerprint generation relied on handcrafted techniques and statistical models for minutiae and orientation field synthesis, often finalized using Gabor filtering methods~\cite{ansari2011generation, Cappelli2004SFinGeA, zhao2012fingerprint, johnson2013texture}. However, these methods introduced issues such as additive noise in the master fingerprint images~\cite{cao2018fingerprint} and failed to preserve realistic fingerprint patterns~\cite{shukla2024vikriti}. In the realm of synthetic fingerprint generation, Cappelli et al. \cite{cappelli2002synthetic} employed Gabor filtering for ridge structure, a random model for minutiae modeling, and the Zero Pole approach for alignment, generating 10,000 plain fingerprints.

With the advent of Generative Adversarial Networks (GANs) \cite{goodfellow2020generative}, recent approaches to synthetic fingerprint generation have shifted from traditional independent statistical methods. Instead, GANs use training data to learn high-dimensional probability distributions, enabling the generation of synthetic fingerprints. However, early GAN techniques, particularly those based on Improved Wasserstein GAN (IWGAN) architecture \cite{gulrajani2017improved}, faced challenges in generating high-resolution images and encountered instability issues.
Finger-GAN~\cite{minaee2018finger} introduced a DCGAN-based framework for generating 64×64 fingerprint images. However, it was evaluated only using the FID metric, lacked conditional control, and did not perform fingerprint matching, uniqueness, or privacy-preservation experiments.


Cao and Jain~\cite{cao2018fingerprint} utilized an improved WGAN (I-WGAN) initialized with a convolutional autoencoder to generate 512 × 512 rolled fingerprint images and demonstrated the ability to synthesize up to 10 million samples efficiently. However, this approach does not generate multiple impressions for a given finger.

Makrushin et al. \cite{makrushin2023privacy} introduced a dataset of synthetic fingerprints, which were generated from modified minutiae templates, with a focus on privacy preservation. Their study illustrated the potential for constructing synthetic fingerprints in a manner that reduces privacy risks while remaining effective for the evaluation of biometric algorithms. Bouzaglo and Keller \cite{bouzaglo2022synthesis} developed a StyleGAN2-based framework for the synthesis and reconstruction of fingerprints and released the SynFing dataset comprising approximately 100,000 image pairs. However, their approach does not consider class-conditional generation by finger number (1–10).

The Clarkson Fingerprint Generator (CFG) \cite{bahmani2021high} employed a progressive growth approach, a GAN-based technique for generating high-fidelity plain impression fingerprints. However, it lacked the capability to generate a fingerprint based on a specific finger (1-10), i.e, conditional fingerprint generation, and was unable to produce multiple impressions of the same fingerprint. 

Mistry et al. \cite{mistry2020fingerprint} utilized an Improved Wasserstein Generative Adversarial Network (I-WGAN) with identity loss to create a large dataset of 100 million synthetic fingerprints, primarily to assess fingerprint search performance at a large scale. However, this approach also does not generate multiple impressions for a given finger.
PrintsGAN \cite{engelsma2022printsgan} is a two-stage fingerprint synthesis pipeline designed to generate realistic and diverse fingerprint images, supporting multiple impressions per identity. However, it lacked spoof generation and conditioning on finger classes. FPGAN-Control~\cite{shoshan2024fpgan} further enabled control over impression-level factors such as fingerprint type (e.g., rolled or slap), scanner type, moisture, and pressure, though it also did not support finger-specific conditioning.

DiffFinger~\cite{grabovski2024difffinger} and GenPrint~\cite{grosz2024universal} leveraged diffusion models for high-quality, diverse generation. While GenPrint allows multimodal prompts (e.g., text, style embeddings) to control acquisition style (e.g., slap, latent), neither approach enables per-finger or spoof material conditioning.

CFG V2~\cite{bahmani2022deep} improved upon the original CFG by introducing conditioning on finger types and age groups, but still did not allow multi-impression generation per class.

SpoofGAN \cite{grosz2022spoofgan} produces both synthetic live and spoof fingerprint images to enhance spoof detection by expanding limited spoof datasets, addressing the issue of insufficient data for training spoof detection algorithms. However, it does not generate synthetic live fingerprint and synthetic spoof fingerprint based on specific finger types.
 
In response to these limitations, our proposed model leverages the architectures of StyleGAN2-ADA \cite{karras2020training}, StyleGAN3 \cite{karras2021alias}, and CycleGAN \cite{zhu2017unpaired} to \emph{generate both synthetic live and spoof fingerprint, conditioned on specific fingers/classes (1–10).} The model also supports the \emph{generation of multiple impressions} of the same fingerprint. This innovative approach addresses prior shortcomings by introducing conditional GAN capabilities tailored for generating realistic and finger specific synthetic fingerprint datasets.
\section{Methods}
In this section, we provide an overview of our proposed architecture for generating synthetic live and spoof fingerprint images for any specific finger/class (1-10), as illustrated in Figure~\ref{fig:overview}. The process begins with two separately trained models StyleGAN2-ADA and StyleGAN3, each capable of generating synthetic live fingerprints conditioned on the selected finger class, as described in Section ~\ref{subsubsec:stylegan2_training} and Section ~\ref{subsubsec:stylegan3_training}. \par
\begin{figure}
 \centering
 \includegraphics[width=1\linewidth]{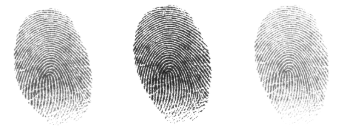} 
 \caption{Three impressions of the same finger from our DB3 synthetic dataset.}
 \label{fig:imp}
\end{figure}

To simulate multiple impressions of the same fingerpint, transformation and deformation techniques were applied as described in Section ~\ref{subsec:GeneratingMultipleImpressions}. In both our approach and SpoofGAN \cite{grosz2022spoofgan}, random transformations such as translation and rotation are applied to simulate different positioning of the fingerprint during capture. However, the key difference lies in how we handle non-linear deformations. SpoofGAN uses a learned statistical deformation model, specifically a dataset of fingerprint videos where minutiae locations were labeled and deformation patterns were extracted using Principal Component Analysis (PCA). In contrast, our approach uses Radial Basis Functions (RBF) to introduce non-linear deformations. These deformations are achieved by randomly placing control points on the fingerprint and applying displacements, creating a smooth warping effect that replicates the elastic nature of the skin. \autoref{fig:imp} shows three multiple impressions of the same finger from our DB3 synthetic dataset.\par 
To generate synthetic spoof fingerprints, the synthetic live fingerprint are further passed through one of eight CycleGAN models as shown in Figure~\ref{fig:overview}, each trained to replicate the visual characteristics of a specific spoof material, including EcoFlex, PlayDoh, Wood Glue, Gelatine, Latex, OOMOO, Silicone, and Body Double.
The entire process is user-configurable:
\begin{itemize}
    \item Selection of the desired finger class
    \item Selection of spoof material (or \textit{live})
    \item Specification of the number of impressions per fingerprint
    \item Specification of the total number of fingerprints to generate
\end{itemize}


\subsection{Training}
We trained two separate models from scratch using the StyleGAN2-ADA and StyleGAN3 architectures. Both models were trained on the same bona fide fingerprint dataset, referred to as DB1, which comprises 20,844 contact-based fingerprints. These images were collected from 338 unique individuals using a Crossmatch Guardian scanner and are part of a private dataset from Clarkson University. 

Each image in the original dataset contains four fingers and two thumbs, with corresponding XML annotation files specifying bounding boxes for each fingertip. Using these annotations, we extracted square regions surrounding each fingertip, resulting in a total of 20,844 individual finger images. All images were resized to 512$\times$512 pixels to ensure compatibility with both StyleGAN2-ADA and StyleGAN3. In addition, a JSON file was generated to label each image with its corresponding finger class (1–10), enabling classification into ten distinct finger categories.

We selected StyleGAN2-ADA and StyleGAN3 due to their proven efficacy in high-resolution image synthesis, particularly in scenarios involving limited training data, which is a common challenge in biometric datasets. StyleGAN2-ADA incorporates adaptive discriminator augmentation (ADA), which mitigates overfitting and improves generalization by dynamically adjusting augmentations during training. This property is especially beneficial when working with relatively small biometric datasets.

StyleGAN3, on the other hand, introduces an alias-free architecture that eliminates spatial aliasing during image generation process, resulting in smooth and continuous ridge patterns and improved spatial coherence in the synthesized images. Its architecture also supports translation equivariance, contributing to stable ridge flow and consistent spatial structure, which are essential properties for generating realistic fingerprint images.

Previous studies have shown that both architectures outperform earlier GAN models in terms of visual fidelity and data efficiency \cite{karras2020training, karras2021alias}, making them particularly suitable for generating diverse, high-quality, and identity-preserving synthetic fingerprint datasets. 

We conducted training on two different hardware platforms: an NVIDIA A100 GPU provided by the NSF Chameleon testbed, and an NVIDIA GeForce RTX 3090 Ti GPU. Detailed training configurations and results for each model are presented in the subsequent sections.

\subsubsection{StyleGAN2-ADA Training}
\label{subsubsec:stylegan2_training}

We trained the StyleGAN2-ADA model from scratch using the DB1 dataset consisting of 20,844 contact-based fingerprint images across ten finger classes (1–10). The model was trained at a resolution of 512×512 using the default non-saturating logistic loss with R1 regularization for the discriminator and path length regularization for the generator. Training was performed for 4000 kimgs (i.e., 4 million images shown to the discriminator), which is estimated to take approximately 6 days on a single GPU. The best-performing model was selected based on the lowest Fréchet Inception Distance (FID) achieved during training, with a minimum FID score of 25.

\subsubsection{StyleGAN3 Training}
\label{subsubsec:stylegan3_training}

We trained the StyleGAN3 model from scratch using the DB1 dataset, which includes 20,844 contact-based fingerprint images categorized into ten finger classes (1–10). Training was conducted at a resolution of 512×512 using the default loss functions proposed in StyleGAN3, including non-saturating logistic loss, R1 regularization for the discriminator, and path length regularization for the generator. The training was performed for 4000 kimgs (i.e., 4 million images shown to the discriminator), with an estimated total runtime of approximately 8.2 days on a single GPU. The best model checkpoint was selected based on the lowest Fréchet Inception Distance (FID) achieved during training, which was 5.

\subsubsection{CycleGAN Training for Spoof Generation}

We trained eight separate CycleGAN models to perform unpaired image-to-image translation from synthetic live fingerprints (domain A) to corresponding spoof fingerprints (domain B). The generator G: A→B was trained to convert live fingerprints into spoofed versions, while the reverse generator F: B→A enforced a cycle-consistency constraint by reconstructing the original live images from their spoofed versions. Each CycleGAN model was dedicated to a specific spoof material: EcoFlex, Play-Doh, Wood Glue, Gelatine, Latex, OOMOO, Silicone, and Body Double. The goal was to generate high-quality synthetic spoofs that visually resemble material-specific presentation attacks, thereby supporting the development and evaluation of robust fingerprint spoof detection systems.

CycleGAN was selected due to its ability to perform unpaired image-to-image translation, making it ideal for scenarios where one-to-one correspondence between source (live) and target (spoof) samples is unavailable. It effectively learns to generate spoof-like textures while preserving biometric features such as ridge flow and structural fidelity through its cycle-consistency mechanism. 

Training was conducted at a resolution of 512$\times$512 using the standard CycleGAN architecture with Least Squares GAN (LSGAN) loss for improved convergence stability. Each model was trained using a combination of adversarial loss, cycle-consistency loss (weighted by $\lambda = 10.0$), and identity loss (weighted by $\lambda_{\text{id}} = 0.5$). We trained each model for 200 epochs (100 constant + 100 linear decay) with a batch size of 2. On a single GPU, training each model took approximately 24 hours, totaling nearly 8 days for all eight spoof-specific generators.

The overall training objective is defined as follows, following the formulation by Zhu \textit{et al.}~\cite{zhu2017unpaired}:

\begin{minipage}{0.5\textwidth}
\begin{equation}
\begin{split}
L(G, F, D_A, D_B) &= L_{\mathrm{GAN}}(G, D_B, A, B) \\
&+ L_{\mathrm{GAN}}(F, D_A, B, A) + \lambda L_{\mathrm{cyc}}(G, F)
\end{split}
\label{eq:cycle_loss}
\end{equation}
\end{minipage}

\noindent

Here, $G: A \rightarrow B$ and $F: B \rightarrow A$ are the generator networks, while $D_A$ and $D_B$ are the corresponding discriminators. 
$L_{\mathrm{GAN}}$ and $L_{\mathrm{cyc}}$ represent the adversarial and cycle-consistency losses, respectively. 
An additional identity loss was employed during training to preserve color and domain-specific features when processing images that already belong to the target domain.

All fingerprint images used for training were sourced from publicly available datasets captured using the CrossMatch fingerprint scanner, specifically from the LivDet 2009~\cite{marcialis2009first}, 2013~\cite{ghiani2013livdet}, and 2015~\cite{mura2015livdet} datasets. To ensure consistency and compatibility with the CycleGAN architecture, all fingerprint images were center-cropped and resized to 512$\times$512 pixels. The architecture modified to support input images at the specified resolution.

Each spoof dataset was carefully balanced by combining samples from LivDet 2009, 2013, and 2015, ensuring an equal number of live and spoof fingerprints. Table~\ref{tab:spoof_dataset_sizes} summarizes the dataset sizes used for each spoof material.
\begin{table}[ht]
\centering
\small
\renewcommand{\arraystretch}{1.2}
\setlength{\tabcolsep}{2pt}
\caption{Dataset Sizes for Each Spoof Material}
\begin{tabular}{|l|c|c|}
\hline
\textbf{Material} & \textbf{Live Fingerprints} & \textbf{Spoof Fingerprints} \\
\hline
Body Double    & 1,095 & 1,095 \\
EcoFlex        &   748 &   748 \\
Gelatine       & 1,600 & 1,600 \\
Latex          &   480 &   480 \\
Wood Glue      &   480 &   480 \\
OOMOO          &   297 &   297 \\
PlayDoh        & 2,417 & 2,417 \\
Silicone       & 1,190 & 1,190 \\
\hline
\end{tabular}
\label{tab:spoof_dataset_sizes}
\end{table}

\subsection{Generating Multiple Impressions}
\label{subsec:GeneratingMultipleImpressions}
To generate multiple impressions of the same finger, the following steps are performed:

\begin{enumerate}
    \item \textbf{Transformation:}
    \begin{itemize}
        \item Translation shifts the fingerprint image by a random number of pixels within the range of [-10, 10] pixels in any direction.
        \item Rotation spins the image around its center by a random angle within the range of [-30°, 30°].
    \end{itemize}
    These transformations simulate varied positions and orientations of the fingerprint on the scanner.

    \item \textbf{Deformation:}
    \begin{itemize}
        \item Non-linear changes are introduced to simulate the elastic nature of the skin.
        \item Radial basis functions are used to create a smooth, non-linear displacement field.
        \item The deformation distorts the image locally in a realistic manner, mimicking how a fingerprint warps under different pressures.
    \end{itemize}
    
    \item \textbf{Mask Creation:}
    \begin{itemize}
        \item A binary mask is created to identify the fingerprint area.
        \item A threshold of 180 is applied, where pixels above this value are considered part of the fingerprint.
    \end{itemize}
    
    \item \textbf{Contrast and Brightness Adjustment:}
    \begin{itemize}
        \item Contrast is varied with an alpha parameter randomly selected between 0.7 and 1.3.
        \item Brightness is adjusted using a beta parameter randomly selected between -30 and 30.
        \item These adjustments are confined to the fingerprint area, preserving the white background and simulating natural variations caused by ink or skin contact.
    \end{itemize}
\end{enumerate}

\section {Results and discussions}
To the best of our knowledge, no existing model has demonstrated the ability to generate both synthetic live and synthetic spoof fingerprint images conditioned on specific fingers (1–10). We compared our approach with the state-of-the-art SpoofGAN \cite{grosz2022spoofgan}, which utilized 1,500 synthetic live fingerprints in their experiments. Unlike SpoofGAN, which does not support class/finger specific fingerprint generation, our method enables conditional generation for each individual finger class (1–10).
\begin{figure}
 \centering
 \includegraphics[width=1\linewidth]{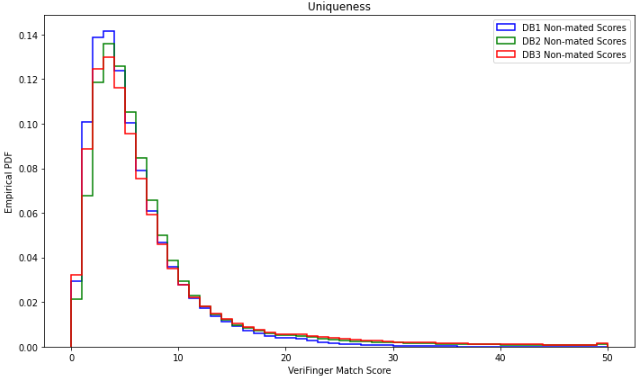} 
 \caption{The VeriFinger matcher analyzed imposter score distributions. Three experiments evaluated uniqueness: DB1 Non-mated scores, DB2 Non-mated scores, and DB3 Non-mated scores. The resulting distributions of match scores (DB2 Non-mated and DB3 Non-mated) closely resemble the distribution of imposter scores for the training images (DB1 Non-mated scores). This indicates that the fingerprints in the generated images exhibit a similar level of uniqueness to those in the training images.}
 \label{fig:uniq.png}
\end{figure}
\subsection{Synthetic Live and Spoof Fingerprint Generation}

To compare our fingerprint generation method with existing methods, we followed the SpoofGAN dataset protocol by generating 1,500 synthetic live fingerprint images using both of our trained models: StyleGAN2-ADA and StyleGAN3. Unlike SpoofGAN, which does not support class-specific generation, our method enables conditional generation for each individual finger class (1–10).

Using the trained StyleGAN2-ADA model and CycleGAN model, we generated 50 synthetic live fingerprints for each of the ten fingers/classes. To simulate intra-class variability, three impressions were generated for each fingerprints, following the procedure outlined in Section~\ref{subsec:GeneratingMultipleImpressions}. This resulted in a total of 1,500 synthetic live fingerprints, and we also generated corresponding 1500 synthetic spoof fingerprints of 8 different spoof materials by using our trained 8 CycleGAN-based models, forming the DB2 dataset. The synthetically generated live fingerprints from DB2 are illustrated in Figure~\ref{fig:gan2}.

Similarly, using the trained StyleGAN3 model and CycleGAN model, we generated another set of 50 synthetic fingerprint images per finger/class, again producing three impressions per finger as describe in Section~\ref{subsec:GeneratingMultipleImpressions}. In total, we produced 1,500 synthetic live fingerprints and 1,500 synthetic spoof fingerprints across 8 different spoof materials by using our trained 8 CycleGAN-based models, forming the DB3 dataset. Synthetic live fingerprints of DB3 are shown in Figure~\ref{fig:gan3}.

\subsection{Feature Similarity Between Bona Fide and Synthetic Fingerprints}
Table~\ref{tab:table1} compares the biometric features of ten fingers from the bona fide dataset (DB1), the synthetic fingerprint datasets (DB2 and DB3), and the SpoofGAN 1500 live fingerprint dataset. 
The fingerprint metrics in the table include Ridge Ending Minutiae Count, Bifurcation Minutiae Count, Reliability of Ridge Minutiae, Reliability of Bifurcation Minutiae, Percentage of Bifurcation Minutiae, Area of the Fingerprint, and NFIQ2 score. These metrics are crucial for evaluating the biometric realism of the generated synthetic fingerprints. We used NIST NFIQ 2.0 \cite{nfiq20160} for the quality metric of the generated and training fingerprint datasets, and MINDTCT from NIST Biometric Image Software (NBIS) \cite{tabassi2004nist} for minutiae quality analysis. 

For the use of synthetic fingerprint data instead of bona fide data, the biometric features of bona fide and synthetic datasets need to be closely aligned \cite{grosz2022spoofgan}. As shown in \autoref{tab:table1}, the ridge ending minutiae count is nearly consistent across DB1 (50.11), DB2 (49.73), and DB3 (49.24), whereas SpoofGAN shows a lower average of 38.45. A similar trend is seen for the bifurcation minutiae count, where the DB1 dataset has a mean of 23.81, DB2 and DB3 show slightly lower values (20.43 and 18.44, respectively), and SpoofGAN is comparable at 19.36.

The reliability scores for both ridge and bifurcation minutiae remain consistent across all datasets, with only minor fluctuations. Notably, DB2 and DB3 have reliability scores similar to the DB1 dataset, whereas SpoofGAN’s reliability is slightly lower but still within a reasonable range.

In terms of the percentage of bifurcation minutiae, DB3 and DB2 are slightly lower (25.91 and 28.06, respectively) compared to the DB1 dataset (31.03), while SpoofGAN exceeds the DB1 dataset at 32.23. The area of the fingerprint is also slightly reduced in DB2 (31.77) and SpoofGAN (26.54), whereas DB3 (34.98) remains relatively close to DB1 (37.10).

Finally, the NFIQ2 score, which indicates overall fingerprint image quality, is highest for DB2 (61.80), followed by DB3 (60.46), both exceeding the DB1 dataset (53.66), whereas the score reported in the SpoofGAN paper is significantly lower at 44.34.

Overall, the comparison confirms that DB2 and DB3 \textit{closely mimic the biometric characteristics of real fingerprints}, with higher quality scores and similar minutiae distributions, thereby supporting their use as effective synthetic substitutes for bona fide fingerprint datasets. \par
Overall, the comparison indicates that the biometric features of both synthetic fingerprint datasets (DB2 and DB3) closely resemble those of the bona fide (DB1) dataset.
\begin{table*}[ht]
\caption{This table presents a comparison of the biometric features of all ten fingers from the bona fide (DB1) dataset, our synthetic fingerprint datasets (DB2 and DB3), and the SpoofGAN \cite{grosz2022spoofgan} dataset.}
\centering
\small
\setlength{\tabcolsep}{2pt}
\begin{tabular}{|l|c|c|c|c|c|c|c|c|}
\hline
\textbf{Measure} & \multicolumn{2}{c|}{\parbox{2.7cm}{\centering \textbf{DB1}}} & \multicolumn{2}{c|}{\parbox{2.7cm}{\centering \textbf{DB2}}} & \multicolumn{2}{c|}{\parbox{2.7cm}{\centering \textbf{DB3}}} & \multicolumn{2}{c|}{\parbox{2.7cm}{\centering \textbf{SpoofGAN}}} \\
\hline
& \textbf{Mean} & \textbf{STD} & \textbf{Mean} & \textbf{STD} & \textbf{Mean} & \textbf{STD} & \textbf{Mean} & \textbf{STD} \\
\hline
Ridge Ending Minutiae Count & 50.11 & 17.59 & 49.73 & 14.32 & 49.24 & 16.22 & 38.45 & 9.95 \\
\hline
Bifurcation Minutiae Count & 23.81 & 14.78 & 20.43 & 12.57 & 18.44 & 13.13 & 19.36 & 10.40 \\
\hline
Reliability of Ridge Minutiae & 0.40 & 0.09 & 0.39 & 0.07 & 0.39 & 0.08 & 0.39 & 0.06 \\ 
\hline
Reliability of Bifurcation Minutiae & 0.53 & 0.15 & 0.59 & 0.15 & 0.58 & 0.16 & 0.55 & 0.12 \\ 
\hline
Percentage of Bifurcation Minutiae & 31.03 & 12.66 & 28.06 & 14.63 & 25.91 & 14.07 & 32.23 & 11.76
\\ 
\hline
Area of the Fingerprint & 37.10 & 11.78 & 31.77 & 8.29 & 34.98&11.53 & 26.54 & 5.97 \\ 
\hline
NFIQ2 Score & 53.66 & 24.44 & 61.80 & 18.12 & 60.46 & 19.36 & 44.34 & 14.38 \\
\hline
\end{tabular}
\label{tab:table1}
\end{table*}

\begin{table}[!t]
\centering
\footnotesize
\caption{True Acceptance Rate (TAR) and False Acceptance Rate (FAR) for bona fide and synthetic datasets. Results are shown at the fixed VeriFinger threshold of 48, and 
at dataset-specific
thresholds corresponding to FAR = 0.01\%.}
\label{tab:TAR}
\renewcommand{\arraystretch}{1.15}
\setlength{\tabcolsep}{2.5pt}
\begin{tabular}{l l r r}
\hline
\textbf{Threshold} & \textbf{Dataset} & \textbf{TAR (\%)} & \textbf{FAR (\%)} \\
\hline
\multirow{4}{*}{48}  & DB1 (Bona fide)        & 94.47 & 0.0057 \\
                     & DB2 (StyleGAN2-ADA)    & 99.60 & 9.2181 \\
                     & DB3 (StyleGAN3)        & 99.47 & 1.4457 \\
                     & SpoofGAN               & 99.40 & 0.2129 \\
\hline
\multirow{3}{*}{75}  & DB2 (StyleGAN2-ADA)    & 99.60 & 7.6709 \\
                     & DB3 (StyleGAN3)        & 99.47 & 0.3272 \\
                     & SpoofGAN               & 99.07 & 0.01 \\
\hline
\multirow{2}{*}{132} & DB2 (StyleGAN2-ADA)    & 99.60 & 3.6771 \\
                     & DB3 (StyleGAN3)        & 99.47 & 0.01 \\
\hline
282                  & DB2 (StyleGAN2-ADA)    & 98.67 & 0.01\\
\hline
\end{tabular}
\end{table}

\subsection{Fingerprint Matching Performance}
In our synthetic datasets (DB2 and DB3), mated (genuine) pairs were created by matching fingerprints with the same seed (subject) and finger/class but different impressions, resulting in 1,500 genuine pairs. A total of 1,102,500 non-mated (imposter) pairs were generated by pairing fingerprints from different seeds (subjects), regardless of finger class. For the SpoofGAN dataset, 1,500 mated (genuine) pairs were similarly generated using different impressions of the same subject (seed), while 1,122,750 non-mated (imposter) pairs were created by pairing fingerprints from different subjects(seeds).

\begin{figure}
 \centering
 \includegraphics[width=1\linewidth]{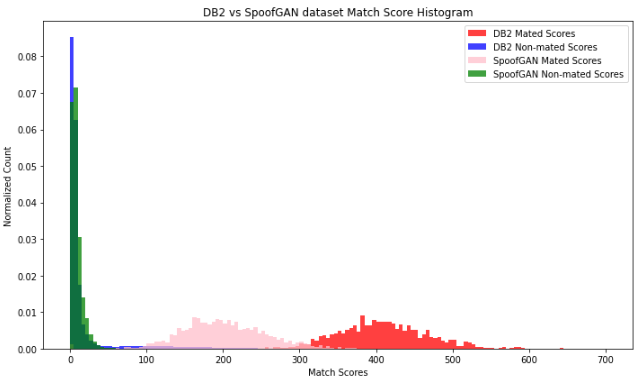} 
 \caption{Comparison of VeriFinger match score distributions between the DB2 and SpoofGAN datasets.}
 \label{fig:db2spoof}
\end{figure}
\begin{figure}
 \centering
 \includegraphics[width=1\linewidth]{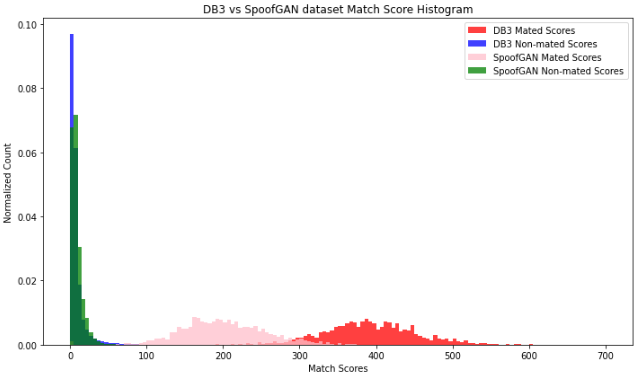} 
 \caption{Comparison of VeriFinger match score distributions between the DB3 and SpoofGAN datasets.}
 \label{fig:db3spoof}
\end{figure}

\begin{figure}
 \centering
 \includegraphics[width=1\linewidth]{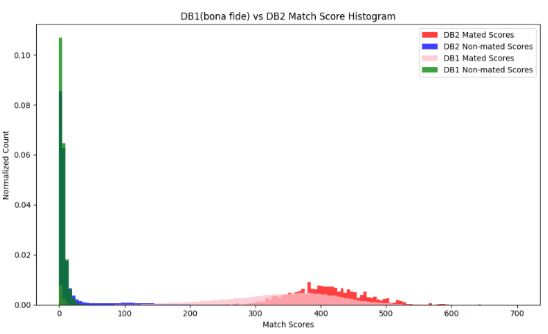} 
 \caption{ The score histogram demonstrates comparable distributions for the DB1 and DB2 datasets, showing similar distributions for both synthetic and bona fide datasets, while maintaining separation between mated and non-mated scores.}

 \label{fig:fig8}
\end{figure}
\begin{figure}
 \centering
 \includegraphics[width=1\linewidth]{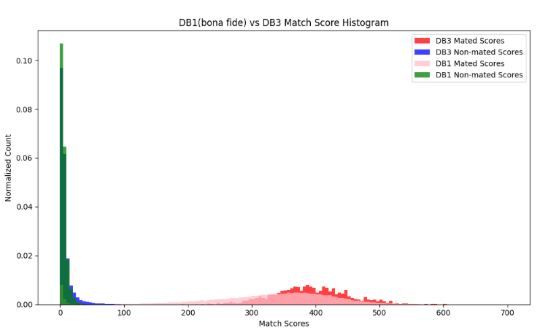} 
 \caption{The score histogram demonstrates comparable distributions for both the DB1 vs. DB3 datasets, showing similar distributions for both synthetic and bona fide datasets, while maintaining clear separation between mated and non-mated scores.}
 \label{fig:fig9}
\end{figure}
Figure~\ref{fig:db2spoof} shows the match score histogram for DB2 vs. SpoofGAN, with strong separation between mated and non-mated scores. Figure~\ref{fig:db3spoof} presents the match score histogram for DB3 vs. SpoofGAN, again with clear separation between mated and non-mated pairs. 
Figure~\ref{fig:fig8} presents the match score histogram for DB1 vs. DB2, illustrating comparable score distributions between the DB2 synthetic dataset and the DB1 bona fide dataset, with a clear separation between mated and non-mated pairs. Similarly, Figure~\ref{fig:fig9} shows the match score histogram for DB1 vs. DB3, where the DB3 synthetic dataset demonstrates similar distributions to the DB1 bona fide dataset and a strong separation between mated and non-mated scores. \par

Table~\ref{tab:TAR} presents TAR at the fixed VeriFinger threshold of 48, as well as at dataset-specific thresholds that achieve 0.01\% FAR. At threshold 48, DB2 achieved the highest TAR of 99.60\%. However, at dataset-specific thresholds for 0.01\% FAR, DB3 achieved the best TAR (99.47\%), followed by SpoofGAN (99.07\%) and DB2 (98.67\%).  

Notably, the VeriFinger thresholds required to achieve 0.01\% FAR in our datasets were significantly higher, 282 for DB2 and 132 for DB3. This indicates that our synthetic fingerprints yield consistently high match scores for genuine comparisons. At dataset-specific thresholds corresponding to FAR = 0.01\%, we obtained TARs of 99.47\% for DB3 and 98.67\% for DB2. These results demonstrate strong discriminability of the synthetic fingerprints.

\subsection{Uniqueness in the Synthetically Generated Fingerprint Datasets}
To evaluate the uniqueness of the generated fingerprints, we analyzed the \textit{imposter (non-mated)} score distributions using three datasets: \textbf{DB1}, the real training dataset consisting of 20,844 live fingerprint images; \textbf{DB2}, a synthetic dataset containing 1,500 live fingerprint images generated using the StyleGAN2-ADA based model; and \textbf{DB3}, a synthetic dataset containing 1,500 live fingerprint images generated using the StyleGAN3 based model.

We conducted three experiments to assess fingerprint uniqueness by evaluating non-mated match scores. The first experiment, DB1 non-mated scores, involved matching 1 million randomly selected non-mated pairs by comparing each fingerprint with those from different subjects. At the VeriFinger-recommended threshold of 48, we observed 57 false matches out of 1 million comparisons.

We then evaluated synthetic datasets under two different thresholds. At dataset-specific thresholds corresponding to FAR = 0.01\%, for DB2, we generated 1,102,500 non-mated pairs by comparing each synthetic fingerprint with those from different subjects/seeds and observed 151 false matches. For DB3, we similarly generated 1,102,500 non-mated pairs, resulting in 142 false matches. 
In contrast, at the fixed threshold of 48, the number of false matches was much higher: 92,181 for DB2 and 14,457 for DB3, reflecting the stricter acceptance levels required to achieve FAR = 0.01\%.

As shown in Figure~\ref{fig:uniq.png}, the resulting non-mated score distributions for DB2 and DB3 closely resemble the bona fide training dataset (DB1). This suggests that the synthetic fingerprints exhibit a comparable level of uniqueness to real fingerprint data (DB1).
  
\subsection{Privacy Preservation}
To evaluate privacy preservation and assess the risk of identity leakage, we conducted fingerprint matching experiments between the bona fide DB1 dataset and the synthetically generated datasets. Specifically, we examined two scenarios: DB1 vs. DB2 and DB1 vs. DB3. 
In the DB1 vs. DB2 case, we evaluated 31.27 million (20,844 × 1,500) comparison pairs, with each pair consisting of one fingerprint image from the DB1 training dataset and one from the DB2 synthetic dataset. None of the comparisons produced a match score exceeding the DB2 VeriFinger threshold for a FAR of 0.01\%. 
Similarly, in the DB1 vs. DB3 case, we evaluated another 31.27 million (20,844 × 1,500) pairs between the DB1 training dataset and the DB3 synthetic dataset. Again, no false matches exceeded the DB3 VeriFinger threshold for a FAR of 0.01\%. At dataset-specific thresholds set to a 0.01\% False Acceptance Rate (FAR), no false matches were detected. However, at the fixed threshold of 48, we identified 118 and 155 false matches for datasets DB2 and DB3, respectively, which correspond to effective FARs of approximately 0.00038\% and 0.00050\%.
These results confirm that the synthetic fingerprints in both DB2 and DB3 exhibit strong privacy preservation with no significant evidence of identity leakage.

\subsection{Impact of Synthetic Fingerprints on Spoof Detection Performance}

To evaluate the utility of our synthetically generated fingerprints in improving spoof detection, we conducted a binary classification experiment using the LivDet 2011 dataset \cite{yambay2012livdet}, which was collected using the DigitalPersona sensor. The dataset contains 2,000 live and 2,000 spoof fingerprint images, where the spoof samples were created using six different materials: Gelatine, Latex, PlayDoh, Silicone, Wood Glue, and Ecoflex. For training, we used 1,500 live and 1,500 spoof images, while the remaining 500 live and 500 spoof images were reserved for testing. All images were resized to 512×512 pixels with white padding to preserve the aspect ratio.

To enhance the training data, we replicated the original LivDet 2011 training set using our synthetic datasets DB2 and DB3. These datasets contain high-quality synthetic live and spoof fingerprints, generated using our StyleGAN-based and CycleGAN-based models. By combining DB2 and DB3 with the original LivDet 2011 training set, we constructed an augmented dataset consisting of 4500 live and 4500 spoof fingerprint images.

We fine-tuned a ResNet50 model \cite{he2016deep}, pretrained on ImageNet, by replacing its final fully connected layer with a binary classifier to distinguish between live and spoof fingerprints. The model was trained for 100 epochs using the cross-entropy loss function and the Adam optimizer. When evaluated on the original LivDet 2011 test set, the model achieved a classification accuracy of 100\%, demonstrating that incorporating synthetic data significantly improved spoof detection performance on previously unseen data.
We also evaluated the model under different training configurations. Training only on the bona fide or real dataset (1,500 live and 1,500 spoof images for training, 500 live and 500 spoof for testing) resulted in an accuracy of 97\%. When trained on synthetic datasets, the accuracy dropped to 52\% on the LivDet 2011 test set, mainly due to the differences in sensors and domains between the synthetic training data, which was based on the Crossmatch scanner, and the test data captured using the DigitalPersona scanner.
Incorporating only 25\% of real data into the synthetic training set raised the accuracy to 81\%, highlighting the importance of real samples for bridging sensor and domain-related differences.
\section{Conclusion}
In this study, we proposed a robust framework for conditional fingerprint generation that supports both synthetic live (bona fide) and synthetic spoof fingerprints. Leveraging the capabilities of StyleGAN2-ADA and StyleGAN3, we developed two high-resolution (512x512) conditional generative models, i.e. models that generate synthetic live fingerprints for specific finger classes (1-10). To create synthetic spoof fingerprints, we employed CycleGANs to translate synthetic live fingerprints into spoof versions across eight different presentation attack materials. We introduced two synthetic datasets, DB2 and DB3. Each dataset contains a total of 1500 synthetic live fingerprints, representing 500 unique fingers (50 images per finger class) with three impressions per finger, along with their corresponding synthetic spoof fingerprints of various materials.

Our StyleGAN3-based model achieved a superior Fréchet Inception Distance (FID) score of 5, indicating high visual realism. Biometric evaluation using NFIQ2 and MINDTCT confirmed that the synthetic fingerprints are highly consistent with real fingerprints in terms of ridge-valley structure, minutiae patterns, and overall quality. 
Further, privacy preservation experiments using VeriFinger fixed threshold confirmed strong privacy preservation, with no significant evidence of identity leakage. The generated datasets also demonstrated a comparable level of uniqueness to the bona fide dataset (DB1), making them suitable for use in biometric research and spoof detection tasks. 

\section{Acknowledgements}
This research was supported by the Center for Identification Technology Research (CITeR). The authors also acknowledge the use of computational resources provided by the Chameleon testbed, supported by the National Science Foundation (NSF).

{\small
\bibliographystyle{ieee_fullname}
\bibliography{main.bib}
}

\end{document}

%% file: regpkg.tex
\usepackage{authblk}
\usepackage{lmodern}
\usepackage[T1]{fontenc}
\usepackage[utf8]{inputenc}
\usepackage[margin=1in]{geometry}
\usepackage{amssymb,amsmath,amsthm,amsfonts}

\usepackage{xcolor}
\usepackage{soul} 
\usepackage{tikz}
\usepackage{pgfplots}
\usepackage{makecell}
\usepackage[inline]{enumitem}
\usepackage{subcaption}
\usepackage{graphicx}
\usepackage{multirow}

\usepackage{pifont}

\usetikzlibrary{arrows,shapes,positioning,shadows,trees}

\tikzset{
  basic/.style  = {draw, text width=2cm, drop shadow, font=\sffamily, rectangle},
  root/.style   = {basic, rounded corners=2pt, thin, align=center,
                   fill=green!30},
  level 2/.style = {basic, rounded corners=6pt, thin,align=center, fill=green!60,
                   text width=8em},
  level 3/.style = {basic, thin, align=left, fill=pink!60, text width=6.5em}
}

\usepackage[pagebackref]{hyperref}
\hypersetup{
    colorlinks,
    linkcolor={red!50!black},
    citecolor={blue!50!black},
    urlcolor={blue!80!black},
    pdfstartview={XYZ null null 1.25}
}

%% file: definitions.tex
\def\mySectionSymbol~{\S{}}

\providecommand{\keywords}[1]
{
  \textbf{\textit{Keywords---}} #1
}